# Complex Matrix Factorization for Face Recognition

Viet-Hang Duong, Yuan-Shan Lee, Bach-Tung Pham, Seksan Mathulaprangsan, Pham The Bao, and Jia-Ching Wang, *Senior Member, IEEE*

*Abstract*— This work proposes a novel method of matrix factorization on the complex domain to obtain both intuitive features and high recognition results in a face recognition system. The real data matrix is transformed into a complex number based on the Euler representation of complex numbers. Base complex matrix factorization (CMF) is developed and two extensions including sparse complex matrix factorization (SpaCMF) and graph complex matrix factorization (GraCMF) are developed by adding sparse and graph constraints. Wirtinger's calculus is used to compute the derivative of the cost function. The gradient descent method is used to solve complex optimization problems. The proposed algorithms are proved to provide effective features for a face recognition model. Experiments on two face recognition scenarios that involve a whole face and an occluded face reveal that the proposed methods of complex matrix factorization provide consistently better recognition results than standard NMFs.

## I. INTRODUCTION

Over the past decade, face recognition has attracted substantial attention [1-6]. In the field of pattern recognition, representing data in a manner that emphasizes relevant information, and transforming a high-dimensional data space into a low-dimensional feature subspace, are important. Subspace methods, such as principal component analysis (PCA) [7], linear discriminant analysis (LDA) [8-9], and nonnegative matrix factorization (NMF) [3], have been successfully used in feature extraction. PCA yields low-dimensional features by projecting data in the directions of largest variance. LDA finds a linear transformation that maximizes discrimination between classes.

Similar to PCA and LDA, which represents data using a linear combination of bases, NMF factorizes the image data into two nonnegative matrices with added non-negativity constraints. Lee and Seung [3] demonstrated that NMF can learn a part-based representation of a face. The variations of NMF was extended by incorporating constraints such as sparsity [10-12], orthogonality [13], discrimination [14], graph regularization [15, 16], and smoothness [15] into the cost function. Nikitidis *et al.* [17] proved that the object function of NMF is non- increasing under projected gradients framework, ensuring the convergence of limit point station. However, as indicated in all of the specified works, most variants of NMF capture only the Euclidean structure of high-dimensional data space and do not consider the nonlinear sub-manifold structure behind the data. Zhang *et al.* [1] provided a solution to this problem that was called topology-preserving non-negative matrix factorization (TPNMF). Unlike the $L_2$ norm, the topology-preserving measure provides the ability to reveal the latent manifold of face patterns. Kernel machines offer an elegant solution by mapping the original data space into a high-dimensional kernel space. Buciu *et al.* [18] presented polynomial non-negative matrix factorization (PNMF), which can be regarded as a nonlinear variation of NMF. In PNMF, images are nonlinearly mapped to polynomial kernel space and then factorized into a nonnegative basis and coefficients. Zafeiriou and Petrou [19] combined the work of [17] and [18] using arbitrary Mercer's kernels [20] to project gradient-based updating without approximating the cost function.

Most of the aforementioned algorithms use the Frobenius norm as a measurement of the reconstruction error. However, the Frobenius norm is well known to be vulnerable to outliers and non-Gaussian noise [21] so the quality of the approximation can be reduced by one or a few corrupted points. To overcome this obstacle, researchers have used several forms of measurements. The $L_{1,2}$ norm was developed to estimate the reconstruction error, based on the assumption of Kong *et al.* that the matching error follows a Laplacian distribution [32].

Liwicki *et al.* [22] recently established the equivalence between the square Frobenius matrix norm in the complex field and the robust dissimilarity measure in the real field. Specifically, they utilized Euler's formula to convert vectors of values of pixel intensity into the unit sphere using Euler's formula. Motivated by this work, our work developed novel complex matrix factorization methods for face recognition; the methods were complex matrix factorization (CMF), sparse complex matrix factorization (SpaCMF), and graph complex matrix factorization (GraCMF). After real-valued data are transformed into a complex field, the complex-valued matrix will be decomposed into two matrices of bases and coefficients, which are derived from solutions to an optimization problem in a complex domain. The generated objective function is the real-valued function of the reconstruction error, which produces a parametric description.

## II. PROPOSED METHOD

### A. Statement of Problems

Let the input data matrix $\mathbf{X} = (\mathbf{x}_1, \mathbf{x}_2, ..., \mathbf{x}_M)$ contains $M$ data vectors as columns. Using the Euler's formula, the elements of real matrix $\mathbf{X}$ are normalized and transformed into a complex number field to yield the complex data matrix $\mathbf{Z}$. An unconstraint optimization problem in an unordered complex field is examined by Problem 1.

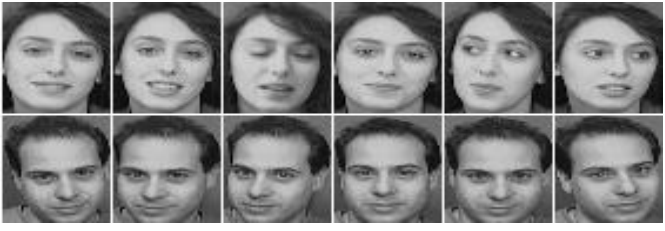

Fig.1. Example of image ORL database [39].

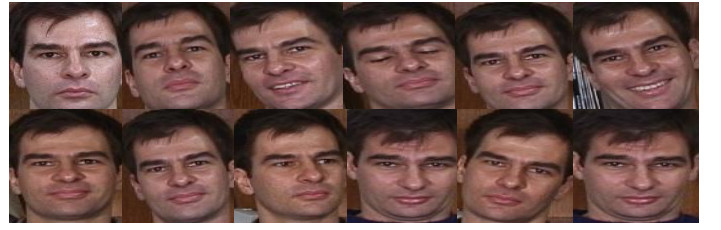

Fig.2. Example of image GT database [46].

**Problem 1**: Given a matrix $\mathbf{Z} \in \mathbb{C}^{N \times M}$, find two matrices $\mathbf{W} \in \mathbb{C}^{N \times K}$ and $\mathbf{V} \in \mathbb{C}^{K \times M}$ that minimize the objective function

$$f(\mathbf{W},\mathbf{V}) = \frac{1}{2}\|\mathbf{Z}-\mathbf{W}\mathbf{V}\|_F^2 \quad (4)$$

where $\|.\|_F$ denotes the Frobenius norm and $K \ll \min(N,M)$.

Additional constraints such as $L_1$ are used to provide sparseness of the coefficient matrix in NMF [10-12]. Likely, a sparsity-constrained matrix factorization on the complex field is developed in Problem 2.

**Problem 2**: Given a matrix $\mathbf{Z} \in \mathbb{C}^{N \times M}$, find two matrices $\mathbf{W} \in \mathbb{C}^{N \times K}$ and $\mathbf{V} \in \mathbb{C}^{K \times M}$ that minimize the objective function

$$f(\mathbf{W},\mathbf{V}) = \frac{1}{2}\|\mathbf{Z}-\mathbf{W}\mathbf{V}\|_F^2 + \alpha \sum_{j=1}^{M}\|\mathbf{V}_{:j}\|_1 \quad (5)$$

where $\sum_{j=1}^{M}\|\mathbf{V}_{:j}\|_1 = \sum_{j=1}^{M}(\sum_{i=1}^{K}|\mathbf{V}_{ij}|)$ and $\alpha$ regulates the balance between the accuracy of the factors and the sparseness of $\mathbf{V}$.

This work proposes another novel model that uses complex matrix factorization; called graph regularized complex matrix factorization (GraCMF). A nearest-neighbor graph $\mathcal{G}$ with $M$ vertices was constructed from the absolute values of $M$ complex data points $(\mathbf{z}_1,\mathbf{z}_2,...,\mathbf{z}_M)$. To define the weight matrix $\mathbf{T}$ on the graph, the cosine similarity is used. As in real domain [16], the complex graph regularization is formulated as,

$$Trace(\mathbf{V}^H \mathbf{L}\mathbf{V}) \quad (7)$$

where $\mathbf{L} = \mathbf{D} - \mathbf{T}$ is the graph of the Laplacian matrix that is induced from the weight matrix $\mathbf{T}$ and a diagonal matrix $\mathbf{D}$, such that $\mathbf{D}_{ii} = \sum_j \mathbf{T}_{ij}$. Incorporating the Laplacian regularize (7) into (4) yields the GraCMF problem as Problem 3.

**Problem 3**: Given a matrix $\mathbf{Z} \in \mathbb{C}^{N \times M}$, find two matrices $\mathbf{W} \in \mathbb{C}^{N \times K}$ and $\mathbf{V} \in \mathbb{C}^{K \times M}$ that minimize the objective function

$$f(\mathbf{W},\mathbf{V}) = \frac{1}{2}\|\mathbf{Z}-\mathbf{W}\mathbf{V}\|_F^2 + \lambda Trace(\mathbf{V}^H \mathbf{L}\mathbf{V}) \quad (8)$$

where $\lambda$ is the regularization parameter.

*B. Complex matrix factorization by gradient descent method*

Equations (4), (5), and (8) are nonconvex minimization problems with respect to both variables $\mathbf{W}$ and $\mathbf{V}$, so finding their optimal solutions is impractical. These NP-hard problems can be solved by using block coordinate descent (BCD) with two matrix blocks [23] to obtain a local solution. The following scheme is utilized herein to solve the three problems in Section III.A.

With $\mathbf{W}$ fixed, the optimization objective functions in (4), (5), and (8) are modified to one variable optimization functions of $\min_{\mathbf{V}} f(\mathbf{W},\mathbf{V})$ and solved by using the gradient descent method [24] with the following update formula

$$\mathbf{V}_{k+1} = \mathbf{V}_k - \beta_k \nabla_{\mathbf{V}_k^*} f(\mathbf{W},\mathbf{V}_k) \quad (26)$$

where $\nabla_{\mathbf{V}^*} f(\mathbf{W},\mathbf{V}_k)$ is the gradient of $f(\mathbf{W},\mathbf{V})$ at $\mathbf{V}_k$, and $\beta_k$ is the step size.

With $\mathbf{V}$ fixed, the update rule for obtaining $\mathbf{W}$ is $\mathbf{W} = \mathbf{V}^\dagger \mathbf{Z}$, where $\dagger$ denotes the Moore–Penrose pseudoinverse.

## III. EXPERIMENTS

Face recognition in the developed models was carried out by firstly computing the pseudoinverse of the basic matrix as $\mathbf{W}_{tr}^\dagger = \mathbf{W}_{tr}^H \left(\mathbf{W}_{tr} \mathbf{W}_{tr}^H\right)^{-1}$. Then, a test sample was encoded as $\mathbf{v}_{te} = \mathbf{W}_{tr}^\dagger \mathbf{z}_{te}$. Finally, one nearest-neighbor (1-NN) classifier was used for recognition.

Extensive experiments were performed on the ORL database [25] and the Georgia Tech database [31] in two scenarios with a whole face and an occluded face. Figures 1, 2, and 3 present unconcluded/occluded faces from the ORL database and the GT database.

To evaluate the performance of the proposed CMF, SpaCMF, and GraCMF, the performance of each is compared with that of the following seven representative algorithms. (1) NMF [3]: standard NMF algorithm; (2) SpaNMF [26]: NMF with sparseness constraint; (3) SpaSemi_NMF [27]: semi-NMF with sparseness constraint; (4) GraNMF [16]: graph regularized NMF; (5) WeNMF [28]: a weighted NMF in which different weights are assigned to reconstruction errors of different entries; (6) MatNMF [29]: uses the Manhattan distance to evaluate error reconstruction; (7) NeNMF [30]: an efficient solver that applies Nesterov's optimal gradient method in optimization process.

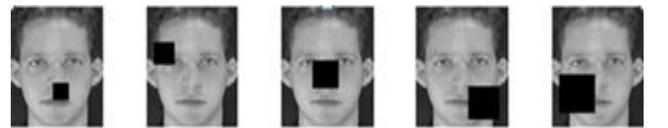

Fig.3. Occluded face samples from ORL dataset with patch sizes of 15 × 15, 20 × 20, 25 × 25, 30 × 30, and 35 × 35, respectively.

TABLE I. BEST FACE-RECOGNITION ACCURACY ACHIEVED USING ORL DATABASE WITH VARIOUS NUMBER OF TRAINING DATA (DIMENSIONS=20)

| No. Trains | CMF | SpaCMF | GraCMF | NMF | SpaNMF | SpaSemi_NMF | GraNMF | WeNMF | MatNMF | NeNMF |
|---|---|---|---|---|---|---|---|---|---|---|
| 5 | **93.65±0.97** | 93.50±1.21 | 89.95±0.48 | 85.02±2.22 | 80.76±2.55 | 49.59±4.12 | 91.83±1.18 | 83.96±2.60 | 89.38±1.45 | 89.97±1.38 |
| 6 | 92.56±1.19 | 92.19±0.87 | **93.63±0.48** | 83.64±2.62 | 81.30±2.65 | 49.97±4.12 | 93.00±0.87 | 83.00±3.24 | 91.95±1.78 | 92.09±1.41 |
| 7 | 94.00±1.35 | 94.25±1.35 | **95.33±0.57** | 82.88±4.70 | 80.58±3.85 | 52.73±5.30 | 93.17±1.00 | 82.62±2.89 | 90.80±2.25 | 93.00±1.38 |
| 8 | 93.63±1.38 | 93.75±1.38 | **97.31±0.46** | 85.08±4.69 | 79.73±4.14 | 53.87±5.17 | 94.94±1.60 | 85.30±3.56 | 94.03±1.93 | 94.13±2.06 |
| 9 | 98.00±1.58 | 99.00±1.58 | **100.0±0.00** | 89.20±4.59 | 82.05±5.59 | 54.35±6.77 | 99.38±1.11 | 87.55±4.24 | 98.05±2.16 | 99.20±1.38 |
| Avg. | 94.37±1.29 | 94.54±0.81 | **95.24±0.40** | 85.16±3.52 | 80.88±3.70 | 52.10±5.10 | 94.46±1.15 | 84.49±3.31 | 92.84±1.91 | 93.68±1.52 |

TABLE II. BEST FACE-RECOGNITION ACCURACY ACHIEVED USING GT DATABASE WITH VARIOUS NUMBER OF TRAINING DATA (DIMENSIONS=20)

| No. Trains | CMF | SpaCMF | GraCMF | NMF | SpaNMF | SpaSemi_NMF | GraNMF | WeNMF | MatNMF | NeNMF |
|---|---|---|---|---|---|---|---|---|---|---|
| 5 | 57.98±1.28 | 58.84±1.22 | **59.60±0.36** | 49.08±2.02 | 46.26±2.27 | 25.94±2.58 | 53.25±1.85 | 48.40±2.29 | 56.21±2.08 | 57.67±1.49 |
| 7 | 65.28±1.60 | 65.18±0.76 | **68.65±0.27** | 51.21±3.02 | 46.31±2.33 | 29.85±2.18 | 57.94±2.78 | 50.65±2.95 | 62.17±2.16 | 63.18±1.76 |
| 9 | 71.57±1.33 | 72.00±1.39 | **77.17±0.77** | 49.81±2.86 | 47.49±2.47 | 28.23±2.31 | 60.10±2.37 | 49.56±3.05 | 63.75±1.72 | 64.82±2.18 |
| 11 | 68.65±1.53 | 69.45±1.69 | **71.00±0.11** | 49.76±3.34 | 48.41±3.45 | 30.92±2.40 | 63.70±3.56 | 49.98±3.55 | 65.6±2.09 | 68.71±1.97 |
| 13 | 76.10±2.28 | 76.10±2.33 | **77.60±0.27** | 50.76±4.49 | 45.00±4.74 | 29.36±3.82 | 65.35±3.75 | 50.64±4.04 | 67.94±2.85 | 70.92±3.35 |
| Avg. | 67.91±1.60 | 68.31±1.48 | **70.80±0.35** | 50.12±3.15 | 46.69±3.05 | 28.86±2.66 | 60.07±2.86 | 49.85±3.18 | 63.13±2.18 | 65.06±2.15 |

*A. Face recognition without occlusion using ORL database*

First, the results obtained using proposed CMF methods were compared with baselines using unoccluded facial images from the ORL database. Five to nine training images of each individual were randomly selected to construct the training set, and the remaining images formed the test set, which was used to estimate the accuracy of face recognition. Table I presents the mean recognition rates; the best results are highlighted in bold. The proposed methods are better than all baselines and GraCMF yielded the best results.

*B. Face recognition using images without occlusion in GT database*

The performances of the proposed methods using the GT database are assessed. The numbers of training images that were used from the GT dataset were 5, 7, 9, 11, and 13. The experimental results thus obtained are presented in Table II. The GT database includes many images that are difficult to recognize so the performances of all methods are lower than those obtained using the ORL database.

*C. Recognition of occluded faces in ORL database*

The images in a test gallery that were cropped to dimensions of 112×92 pixels, occlusion was simulated using sheltering patches of sizes {15×15, 20×20, 25×25, 30×30, 35×35} at random locations. The images were then resized to 28×21. Then, they were randomly selected for training and testing with the ratio of 4:6 and tested several times on sorted test images by occlusion sizes. Table III compares the recognition rates achieved using the three proposed models with the baselines. The outstanding results of CMF, SpaCMF, and GraCMF demonstrate their better ability to handle outliers.

*D. Recognition of occluded faces in GT database*

With occlusions of the same size as those in the images in the ORL database, experiments are performed on occluded faces from the GT database. Table IV presents the important achievements of our methods. The excellent results achieved using GraCMF and GraNMF demonstrate the effectiveness of imposing a geometric constraint on the recognition system. Most importantly, the proposed methods work well not only in recognizing unoccluded faces but also in recognizing occluded faces.

## IV. CONCLUSION

This work developed the new approaches to complex matrix factorization for face recognition. Two additional constraints including sparse penalty and graph regularization are imposed on complex domain. Experimental results reveal that the proposed methods, including CMF, SpaCMF, and GraCMF, yield promising results by extending the core concept of NMF from the real number field to the complex field. We expect that the proposed methods will be stable when applied to other tasks

TABLE III. BEST FACE-RECOGNITION ACCURACY ACHIEVED USING OCCLUDED ORL DATABASE (mean% ± std)

| Occluded | CMF | SpaCMF | GraCMF | NMF | SpaNMF | SpaSemi_NMF | GraNMF | WeNMF | MatNMF | NeNMF |
|---|---|---|---|---|---|---|---|---|---|---|
| 15×15 | 84.88±0.92 | 85.21±1.11 | **86.56±0.96** | 74.32±3.06 | 72.55±2.75 | 45.16±4.04 | 81.25±1.35 | 74.18±2.85 | 80.69±2.37 | 81.32±1.86 |
| 20×20 | 76.42±0.95 | 76.79±1.03 | **78.19±1.04** | 65.45±2.77 | 62.15±3.00 | 41.52±3.02 | 71.23±2.62 | 65.00±2.35 | 72.66±2.20 | 72.95±1.89 |
| 25×25 | 71.38±1.59 | **73.17±1.30** | 72.31±1.77 | 55.18±3.10 | 52.38±3.71 | 35.54±4.23 | 62.19±3.51 | 55.00±2.71 | 64.98±2.82 | 65.39±2.73 |
| 30×30 | 59.42±1.85 | **61.54±1.36** | 61.21±1.47 | 45.62±3.38 | 43.87±6.90 | 28.53±4.15 | 55.21±3.25 | 45.89±3.27 | 56.15±3.58 | 55.58±2.66 |
| 35×35 | 40.54±1.63 | 41.00±1.57 | **42.33±1.73** | 33.63±2.71 | 31.06±3.26 | 23.25±3.53 | 38.79±2.71 | 33.39±2.72 | 40.25±2.89 | 41.33±2.92 |
| Avg. | 66.53±1.39 | 67.54±1.27 | **68.12±1.39** | 54.84±3.00 | 52.40±3.92 | 45.16±4.04 | 61.73±2.69 | 54.69±2.78 | 62.95±2.77 | 63.31±2.41 |

TABLE IV. BEST FACE-RECOGNITION ACCURACY ACHIEVED USING OCCLUDED GT DATABASE (mean% ± std)

| Occluded | CMF | SpaCMF | GraCMF | NMF | SpaNMF | SpaSemi_NMF | GraNMF | WeNMF | MatNMF | NeNMF |
|---|---|---|---|---|---|---|---|---|---|---|
| 15×15 | 63.88±1.32 | 63.94±1.00 | **66.43±0.63** | 48.38±2.48 | 46.48±2.0 | 28.39±1.52 | 54.99±3.51 | 47.79±1.89 | 59.08±2.17 | 59.58±1.58 |
| 20×20 | 62.19±1.23 | 62.23±1.46 | **66.24±0.78** | 46.49±2.18 | 44.51±2.38 | 29.13±2.71 | 56.23±2.27 | 46.49±2.64 | 56.94±2.08 | 57.68±1.95 |
| 25×25 | 60.95±0.92 | 60.85±1.14 | **63.55±0.89** | 43.38±1.92 | 41.81±1.99 | 25.06±1.76 | 51.94±2.83 | 43.83±2.35 | 56.01±1.49 | 56.58±1.24 |
| 30×30 | 53.80±1.06 | 53.66±1.39 | **57.34±0.76** | 37.13±2.47 | 36.55±1.79 | 23.76±2.41 | 47.21±2.33 | 37.53±1.78 | 49.78±1.82 | 49.89±1.84 |
| 35×35 | 52.10±1.17 | 52.65±1.49 | **55.30±0.61** | 35.86±2.34 | 34.69±2.56 | 23.38±2.36 | 46.2±2.02 | 35.14±2.29 | 47.19±2.56 | 48.31±2.25 |
| Avg. | 58.59±1.14 | 58.67±1.31 | **61.77±0.75** | 42.25±2.28 | 40.81±2.14 | 25.94±2.15 | 51.31±2.59 | 42.156±2.19 | 53.8±2.024 | 54.41±1.77 |